\documentclass[11pt,a4paper]{article}
\usepackage[hyperref]{emnlp2020}
\usepackage{times}
\usepackage{latexsym}
\usepackage{amssymb}
\usepackage{amsmath}
\usepackage{bm}
\usepackage{pgfplots}
\usepackage{caption}
\usepackage{subcaption}
\usepackage{graphicx}
\usepackage{xparse}
\usepackage{booktabs}       

\usepackage{microtype}

\aclfinalcopy 
\NewDocumentCommand{\gin}{ mO{} }{\textcolor{orange}{\textsuperscript{\textit{Gin}}\textsf{\textbf{\small[#1]}}}}
\NewDocumentCommand{\jim}{ mO{} }{\textcolor{red}{\textsuperscript{\textit{Jimmy}}\textsf{\textbf{\small[#1]}}}}

\title{Inserting Information Bottlenecks for Attribution in Transformers}

\author{Zhiying Jiang, Raphael Tang, Ji Xin \and Jimmy Lin \\[1ex]
  David R. Cheriton School of Computer Science \\
  University of Waterloo \\[1ex]
  \texttt{\{zhiying.jiang, r33tang, ji.xin, jimmylin\}@uwaterloo.ca}}

\date{}

\begin{document}
\maketitle
\begin{abstract}
Pretrained transformers achieve the state of the art across tasks in natural language processing, motivating researchers to investigate their inner mechanisms.
One common direction is to understand what features are important for prediction.
In this paper, we apply information bottlenecks to analyze the attribution of each feature for prediction on a black-box model.
We use BERT as the example and evaluate our approach both quantitatively and qualitatively.
We show the effectiveness of our method in terms of attribution and the ability to provide insight into how information flows through layers.
We demonstrate that our technique outperforms two competitive methods in degradation tests on four datasets.
Code is available at \url{https://github.com/bazingagin/IBA}.

\end{abstract}
\section{Introduction}

Increasingly prominent is the urge to interpret deep neural networks, with the success of these black-box models remaining vastly inexplicable both theoretically and empirically.
Within natural language processing (NLP), this desire is particularly true for the pretrained transformer, which has witnessed an influx of literature on interpretability analysis.
Such papers include visualizing transformer attention mechanisms~\cite{kovaleva2019revealing}, probing the geometry of transformer representations~\cite{hewitt2019structural}, and explaining the span predictions of question answering models~\cite{van2019does}.

In this paper, we focus on prediction attribution methods.
That is, we ask, ``Which hidden features contribute the most toward a prediction?''
To resolve this question, a number of methods \cite{selvaraju2017grad, smilkov2017smoothgrad} generate attribution scores for features, which provide a human-understandable ``explanation'' of how a particular prediction is made at the instance level.
Specifically, given an instance, these methods assign a numerical score for each hidden feature denoting its relevance toward the prediction.

Previous papers have demonstrated that gradient-based methods fail to capture all the information associated with the correct prediction~\cite{li2016visualizing}.
To address this weakness, \citet{schulz2020iba} insert information bottlenecks~\cite{tishby2000information} for attribution, attaining both stronger empirical performance and a theoretical upper bound on the information used.
Additionally, mutual information is unconstrained by model and task~\cite{guan2019towards}.
Thus, we adopt information bottlenecks for attribution (IBA) to interpret transformer models at the instance level.
We apply IBA to BERT \cite{devlin2019bert} across five datasets in sentiment analysis, textual entailment, and document classification.
We show both qualitatively and quantitatively that the method capably captures information in the model's token-level features, as well as insight into cross-layer behavior.

Our contributions are as follows:~First, we are the first to apply information bottlenecks (IB) for attribution to explain transformers.
Second, we conduct quantitative analysis to investigate the accuracy of our method compared to other interpretability techniques. Finally, we examine the consistency of our method across layers in a case study.
Across four datasets, our technique outperforms integrated gradients (IG) and local interpretable model-agnostic explanations (LIME), two widely adopted prediction attribution approaches.

\section{Related Work}
In terms of scope, interpretability methods can be categorized as model specific or model agnostic. 
Model-specific methods interpret only one family of models, whereas

model-agnostic techniques aim for wide applicability across many families of parametric models.
We can roughly separate model-agnostic methods into three categories:~(1) gradient-based ones~\cite{li2016visualizing, fong2017interpretable, sundararajan2017axiomatic}; (2) probing~\cite{ribeiro2016should, lundberg2017unified,tenney2019you,clark2019does, liu2019linguistic}; (3) information-theoretical methods~\cite{bang2019explaining, guan2019towards, schulz2020iba, pimentel2020information} . 

Gradient-based methods are, however, limited to models with differentiable neural activations.
They also fail to capture all the information associated with the correct prediction~\cite{li2016visualizing}.
Although probing methods provide detailed insight into specific models, they fail to capture inner mechanisms like how information flows through the network~\cite{guan2019towards}. Information-theoretic methods, in contrast, provide consistent and flexible explanations, as we show in this paper.

\citet{guan2019towards} use mutual information to interpret NLP models across different tokens, layers, and neurons, but they lack a quantitative evaluation.
\citet{bang2019explaining} also propose a model-agnostic interpretable model using IB; however, they limit the information through the network by sampling a given number of words at the beginning, which restricts the explanation to neurons only.
Our method is inspired by \citet{schulz2020iba}, who use IBA in image classification. 
\section{Method}

The idea of IBA is to restrict the information flowing through the network for every single instance, such that only the most useful information
is kept.

Concretely, given an input $\mathbf{X}\in\mathbb{R}^N$ and output $\mathbf{Y}\in\mathbb{R}^M$, an information bottleneck is an intermediate representation $\mathbf{T}$ that maximizes the following function:
\begin{equation}
    \operatorname{I}(\mathbf{Y}; \mathbf{T}) - \beta \cdot \operatorname{I}(\mathbf{X}; \mathbf{T}),
\end{equation}
where $\operatorname{I}$ denotes mutual information and $\beta$ controls the trade-off between reconstruction
$\operatorname{I}(\mathbf{Y}; \mathbf{T})$
and information restriction
$\operatorname{I}(\mathbf{X}; \mathbf{T})$.
The larger the $\beta$, the narrower the bottleneck, i.e., less information is allowed to flow through the network. 

We insert the IB after a given layer $l$ in a pretrained deep neural network.
In this case, $\mathbf{X}=f_l(\mathbf{H})$ represents the chosen layer's output, where $\mathbf{H}$ is the input of the layer.
We restrict information flow by injecting noise into the original input:
\begin{equation}
    \mathbf{T} = \boldsymbol{\mu}\odot\mathbf{X} + (\mathbf{1}-\boldsymbol{\mu})\odot\epsilon,
\end{equation}
where $\odot$ denotes element-wise multiplication, $\epsilon$ the injected noise, $\mathbf{X}$ the latent representation of the chosen layer, $\mathbf{1}$ the all-one vector, and $\boldsymbol{\mu}\in\mathbb{R}^N$ the weight balancing signal and noise.
For every dimension $i$, $\boldsymbol{\mu}_i \in [0,1]$, meaning that when $\boldsymbol{\mu}_i=1$, there is no noise injected into the original representation.
To simplify the training process, we set $\boldsymbol{\mu}_i = \sigma(\boldsymbol{\alpha}_i)$,

where $\sigma$ is the sigmoid function and $\boldsymbol{\alpha}$ is a learnable parameter vector. 
In the extreme case, where all the information in $\mathbf{T}$ is replaced with noise ($\mathbf{T} = \epsilon$), it's desirable to keep $\epsilon$ the same mean and variance as $\mathbf{X}$  in order to preserve the magnitude of the input to the following layer. Thus, we have $\epsilon\sim\mathcal{N}(\mu_{\mathbf{X}}, \sigma_{\mathbf{X}}^2)$.

After obtaining $\mathbf{T}$, we evaluate how much information $\mathbf{T}$ still contains about $\mathbf{X}$, which is defined as their mutual information:
\begin{equation}
    \operatorname{I}(\mathbf{X}; \mathbf{T}) = \mathbb{E}_{\mathbf{X}}[D_{KL}[P(\mathbf{T} |\mathbf{X}) \| P(\mathbf{T})]],
\end{equation}
where $D_{KL}$ means Kullback--Leibler (KL) divergence, $P(\mathbf{T}|\mathbf{X})$ and $P(\mathbf{T})$ represent their probability distributions.
While $P(\mathbf{T}|\mathbf{X})$ can be sampled empirically, $P(\mathbf{T})$ 
has no analytical solution since it requires integrating over the feature map
$P(\mathbf{T})=\int P(\mathbf{T}|\mathbf{X})P(\mathbf{X})\mathrm{d}\mathbf{X}$.
As is standard, we use the variational approximation $Q(\mathbf{T}) = \mathcal{N}(\mu_{\mathbf{X}}, \sigma_{\mathbf{X}}^2)$ to substitute $P(\mathbf{T})$, assuming every dimension of $\mathbf{T}$ is independent and normally distributed.
Even though the independence assumption does not hold in general, it only overestimates the mutual information, giving a nice upper bound of mutual information between $\mathbf{X}$ and $\mathbf{T}$:
\begin{subequations}
\begin{align}
    \operatorname{I}(\mathbf{X}; \mathbf{T}) 
                                             & = \mathbb{E}_{\mathbf{X}}[D_{KL}[P(\mathbf{T} |\mathbf{X}) \| Q(\mathbf{T})]] \label{eq:mi} \nonumber \\
                                             & - D_{KL}[Q(\mathbf{T}) \| P(\mathbf{T})] \\
    \operatorname{I}(\mathbf{X}; \mathbf{T}) & 
    \leq \mathbb{E}_{\mathbf{X}}[D_{KL}[P(\mathbf{T} |\mathbf{X}) \| Q(\mathbf{T})]]. \label{eq:ieq}
\end{align}
\end{subequations}
The complete derivation of Equation~\ref{eq:ieq} is in Appendix~\ref{B}.
Since we expect $\operatorname{I}(\mathbf{X},\mathbf{T})$ to be small and mutual information to be always nonnegative, the upper bound is a desired property. 

Intuitively, the purpose of maximizing $\operatorname{I}(\mathbf{Y}; \mathbf{T})$ is to make accurate predictions. Therefore, instead of directly maximizing $\operatorname{I}(\mathbf{Y}; \mathbf{T})$, we minimize the loss function for the original task, e.g., the cross entropy $\mathcal{L}_{\text{CE}}$ for classification problems after inserting the information bottleneck.

\begin{table*}[t]
    \centering
    \begin{tabular}{c|c|c|c|c|c}
        & IMDB & MNLI Matched & MNLI Mismatched & AG News & RTE \\
        \toprule
  Random & 0.011 & 0.106 & 0.106 & 0.008 & 0.012\\
  LIME & 0.038 & 0.244 & 0.260 & 0.033 & 0.014\\
  IG & 0.090 & 0.226 & 0.233 & \textbf{0.036} & 0.043\\
  IBA  & \textbf{0.229} & \textbf{0.374} & \textbf{0.367} & 0.029 & \textbf{0.059}\\
        \bottomrule
    \end{tabular}
    \caption{Absolute probability drop for the target class after the top 11\% most important tokens removed. The larger the score, the more effective the method.}
    \label{tab:raw}
\end{table*}

Combining the above two parts, our final loss function $\mathcal{L}$ is
\begin{equation}
        \mathcal{L} =  \mathcal{L}_{\text{CE}}+ \beta \cdot \mathbb{E}_{\mathbf{X}}[D_{KL}[P(\mathbf{T| \mathbf{X}}) \| Q(\mathbf{T})]].
\end{equation}

Note that we negate the sign for minimization.
The $\beta$ hyperparameter controls the relative importance between the two loss components.
After the optimization process, we obtain for every instance a compressed representation $\mathbf{T}$. 

We then calculate $D_{KL}[P(\mathbf{T} |\mathbf{X}) \| Q(\mathbf{T})]$, indicating how much information is still kept in $\mathbf{T}$ about $\mathbf{X}$, which suggests the contribution of each token and feature.
To generate the attribution map, we sum over the feature--token axis, obtaining the attribution score of each token.

Overall, we try to learn a compressed hidden representation $\mathbf{T}$ that has just enough information about the input $\mathbf{X}$ to predict the output $\mathbf{Y}$.
This compression is done by adding noise, which removes the least relevant feature-level information, with $\boldsymbol{\mu}$ controlling how much to remove.

\section{Experiments}

Through experimentation, we analyze IBA both quantitatively and qualitatively to understand how it interprets deep neural network across layers.

\begin{figure*}[ht]
    \centering
    
    \begin{subfigure}{.29\textwidth}
        \centering
        \includegraphics[width=1\linewidth]{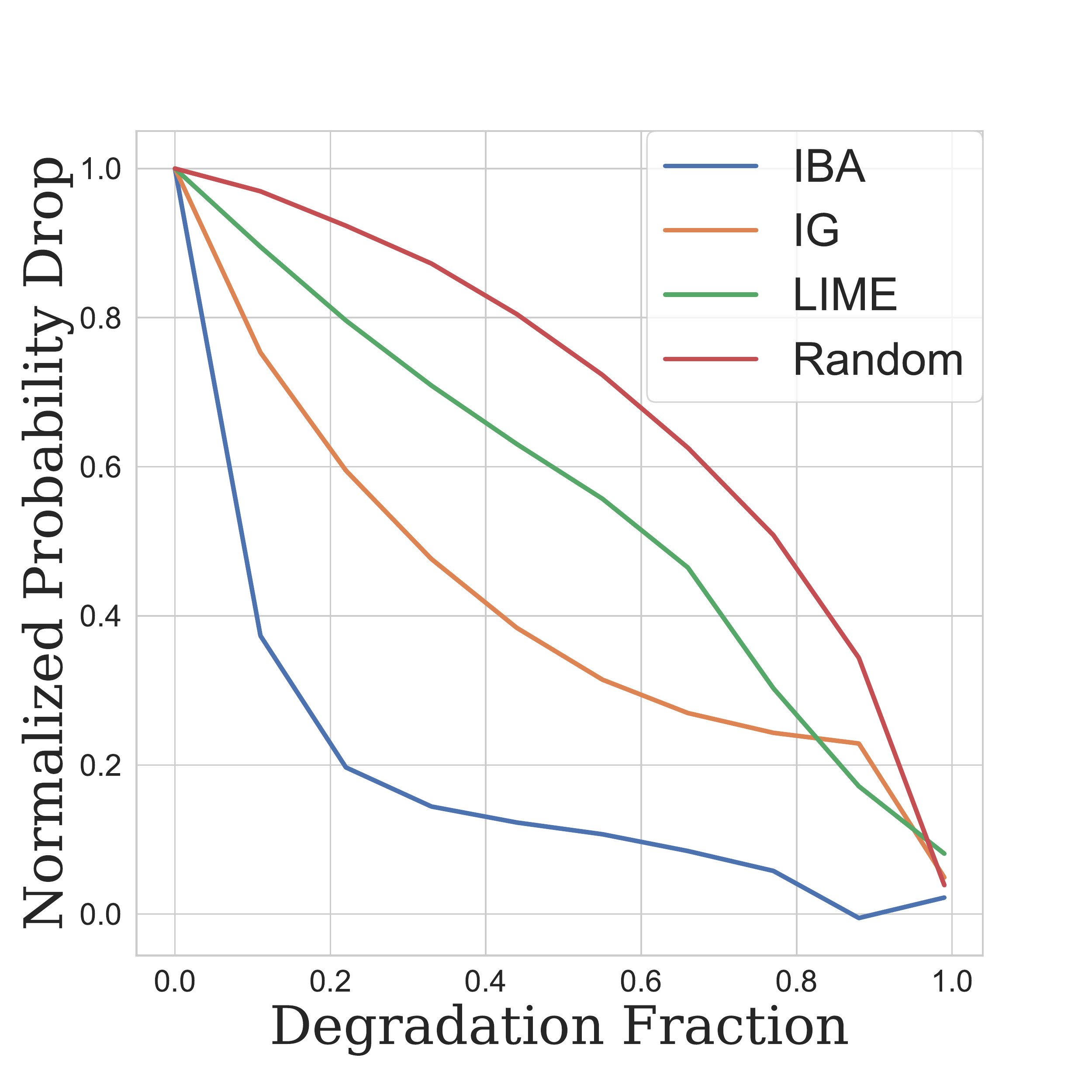}
        \caption{IMDB}
    \end{subfigure}%
    \begin{subfigure}{.29\textwidth}
        \centering
        \includegraphics[width=1\linewidth]{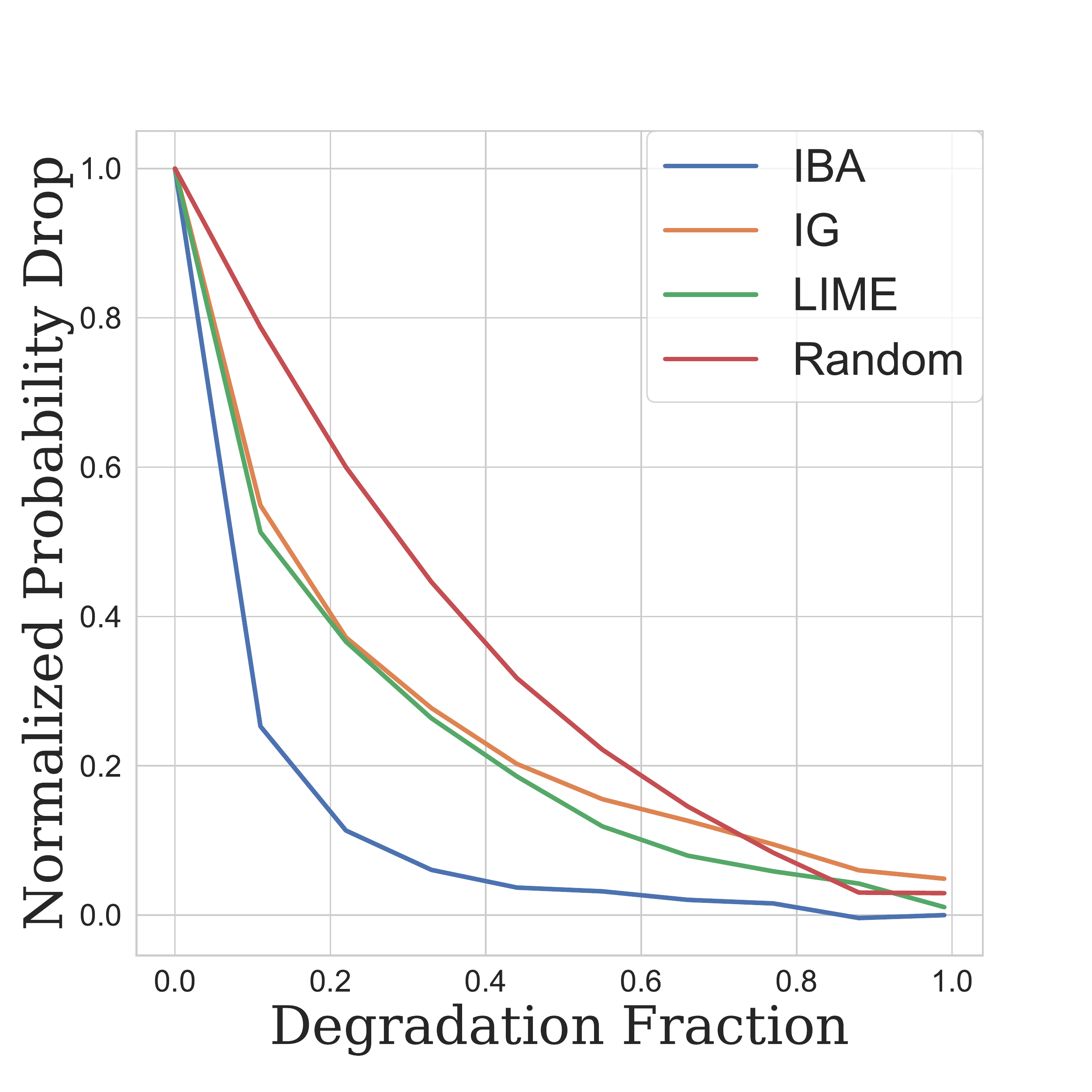}
        \caption{MNLI Matched}
    \end{subfigure}
    \begin{subfigure}{.29\textwidth}
        \centering
        \includegraphics[width=1\linewidth]{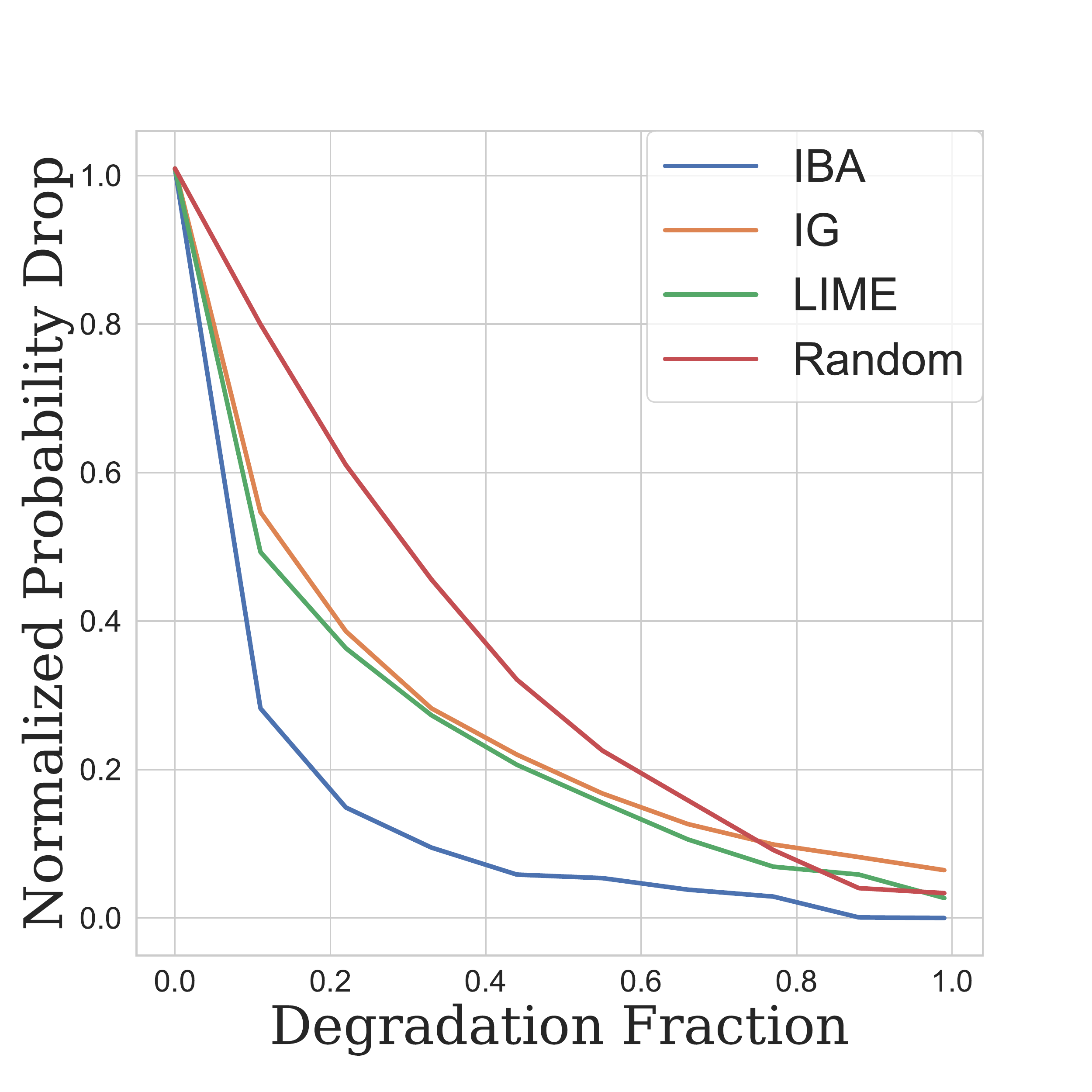}
        \caption{MNLI Mismatched}
    \end{subfigure} %
    \begin{subfigure}{.29\textwidth}
        \centering
        \includegraphics[width=1\linewidth]{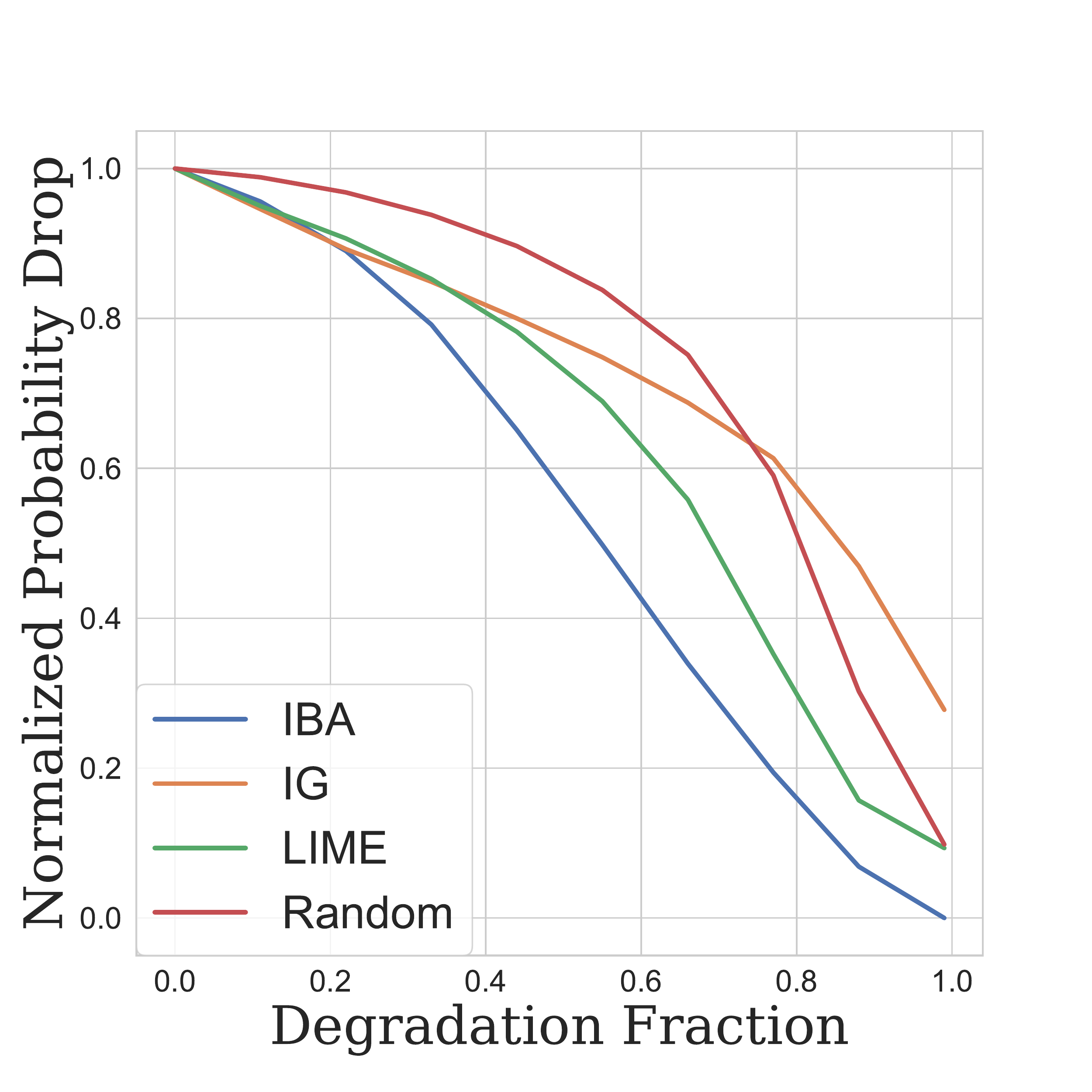}
        \caption{AG News}
    \end{subfigure}
    \begin{subfigure}{.29\textwidth}
        \centering
        \includegraphics[width=1\linewidth]{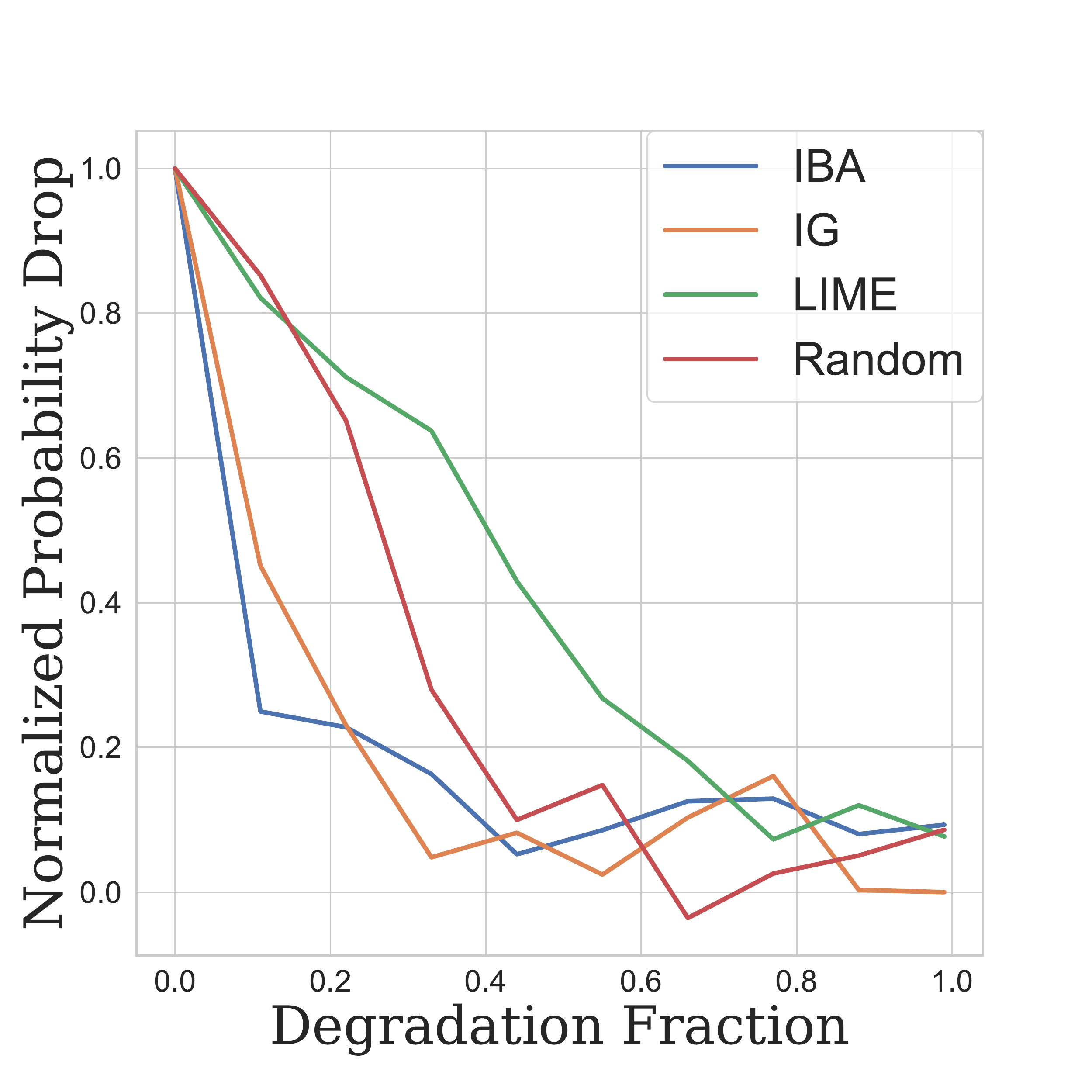}
        \caption{RTE}
    \end{subfigure}%

\caption{Degradation test results comparing IBA, IG, LIME, and random.}
\label{fig:main}

\end{figure*}

\subsection{Experimental Setting}
We compare our method on BERT with two other representative model-agnostic instance-level methods---LIME \cite{ribeiro2016should}, which explores interpretable models for approximation and explanation, and integrated gradients (IG) \cite{sundararajan2017axiomatic}, a variation on computing the gradients of the predicted output with respect to input features.
For a simple baseline, we also compare with ``random,'' whose attribution scores are assigned randomly to tokens.
On each dataset, we fine-tune BERT and apply these interpretability techniques to the model.
We note the test accuracy and generate an attribution score for each token.
Details of all parameters are attached in Appendix~\ref{A}.

There is no consensus on how to evaluate interpretability methods quantitatively \cite{molnar2019}.
LIME's simulated evaluation leverages the ground truth of already interpretable models like decision trees, but the ground truth is unavailable for black-box models like neural networks.
Therefore, we follow \citet{ancona2018towards} and \citet{hooker2018evaluating} and carry out a \textit{degradation test} on IMDB~\cite{maas-EtAl:2011:ACL-HLT2011}, AG News~\cite{gulli_2004}, MNLI~\cite{williams-etal-2018-broad}, and RTE \cite{wang2018glue}, covering sentiment analysis, natural language inference, and text classification.

The degradation test has the following steps: 
\begin{enumerate}
    \item Generate attribution scores $s$ for each interpretability method $f$: $s = f(\mathcal{M}, x, y)$, where $x$ is the test instance, $y$ is the target label, and $\mathcal{M}$ is the model. 
    \item Sort tokens by their attribution score in descending order.
    \item Remove top $k$ tokens to obtain $x'$, the degraded instance; $k$ can be preset.
    \item Test the target class probability $p(y|x')$ with the original model on the degraded instance.
    \item Repeat steps 3 and 4 until all tokens removed.
\end{enumerate}
For the final visualization, we average all test instances at each degradation step to compute $\bar{p}(y|x')$.
Then, we normalize the degradation test result $\bar{p}(y|x')$ to $[0,1]$ using the normalized probability drop $\bar{d} = \frac{\bar{p}(y|x')-m}{o-m}$, where $o$ means the original probability on the nondegraded instance, and $m$ means the minimum of the \textit{fully} degraded instance's probability across all interpretability models.
In this way, the normalized probability drop $\bar{d}$ will be independent of the original model quality and easily comparable across models.
Note that, for IBA, we perform the degradation test on the original model, not the one with the inserted bottleneck.
Thus, a large $\beta$ does not directly cause the probability to drop.
An effective attribution map can find the most important tokens, which means $\bar{p}(y|x')$ after the degradation step will drop substantially.

\subsection{Results and Analysis}
Overall, the results show that our method better identifies the most important tokens compared to other model-agnostic interpretability methods.

\begin{figure*}[t]
    \centering
    \begin{subfigure}{.34\textwidth}
        \centering
        \includegraphics[width=1\linewidth]{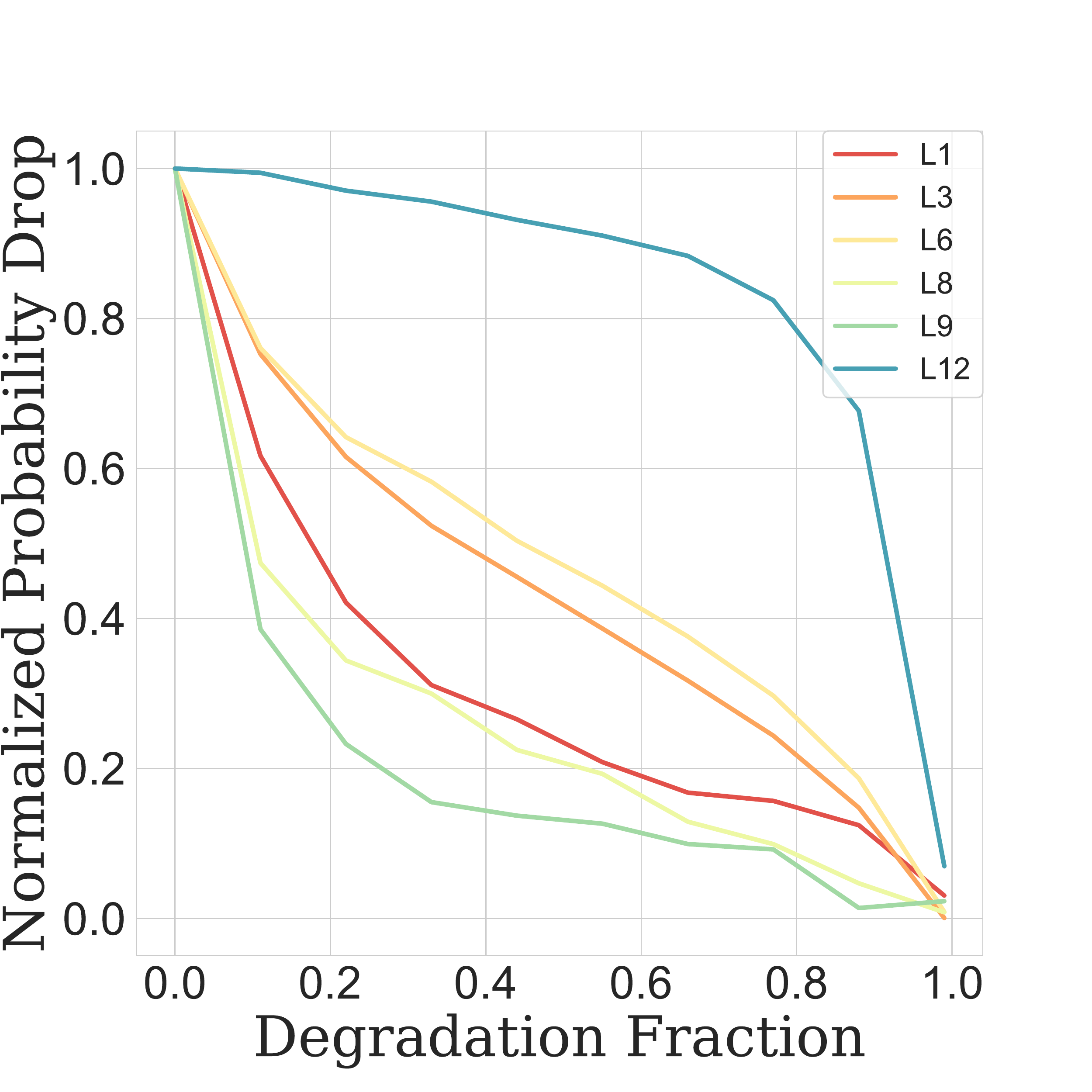}
        \caption{IB after different layers.}
        \label{subfig:layer}
    \end{subfigure}
    \begin{subfigure}{.34\textwidth}
        \centering
        \includegraphics[width=1\linewidth]{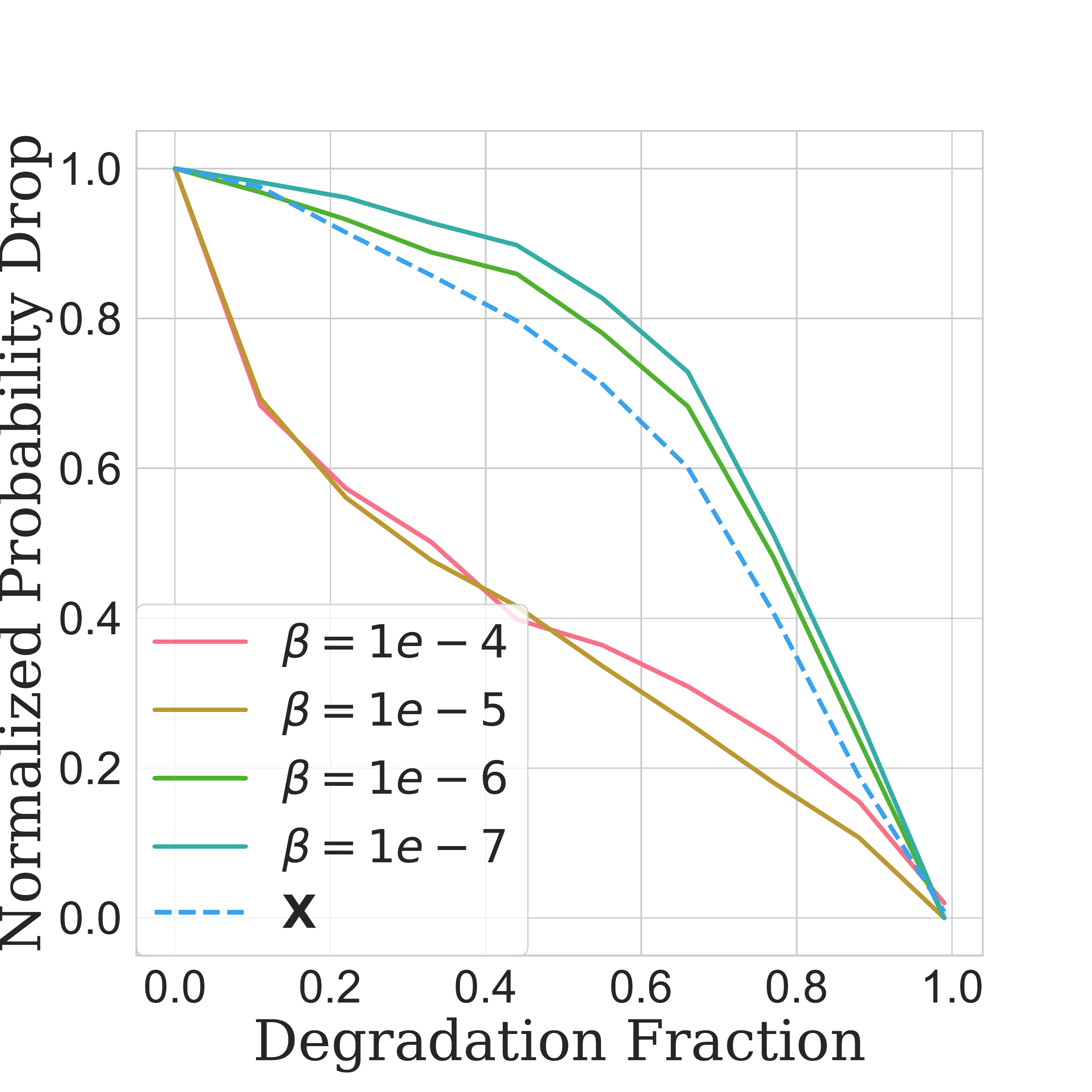}
        \caption{IB with different $\beta$.}
        \label{subfig:beta}
    \end{subfigure}
    \vspace{-0.1cm}
    \caption{Analysis of different layers and different $\beta$.}
    \label{fig:analysis}
    \vspace{0.1cm}
\end{figure*}

\begin{figure*}
    \centering
    \begin{subfigure}{.34\textwidth}
        \centering
        \includegraphics[width=1\linewidth]{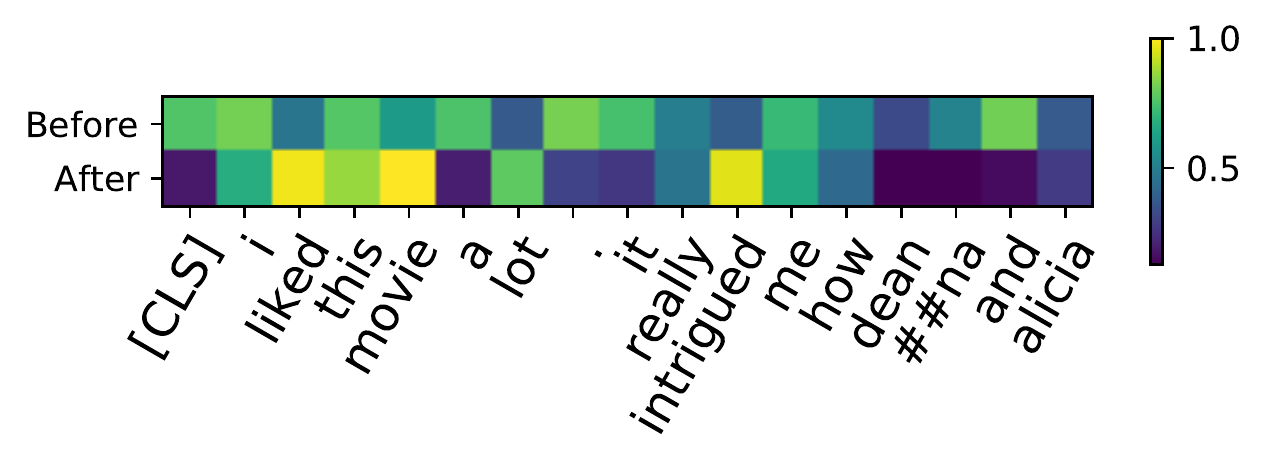}
        \label{subfig:full}
    \end{subfigure}
    \begin{subfigure}{.3\textwidth}
        \centering
        \includegraphics[width=1\linewidth]{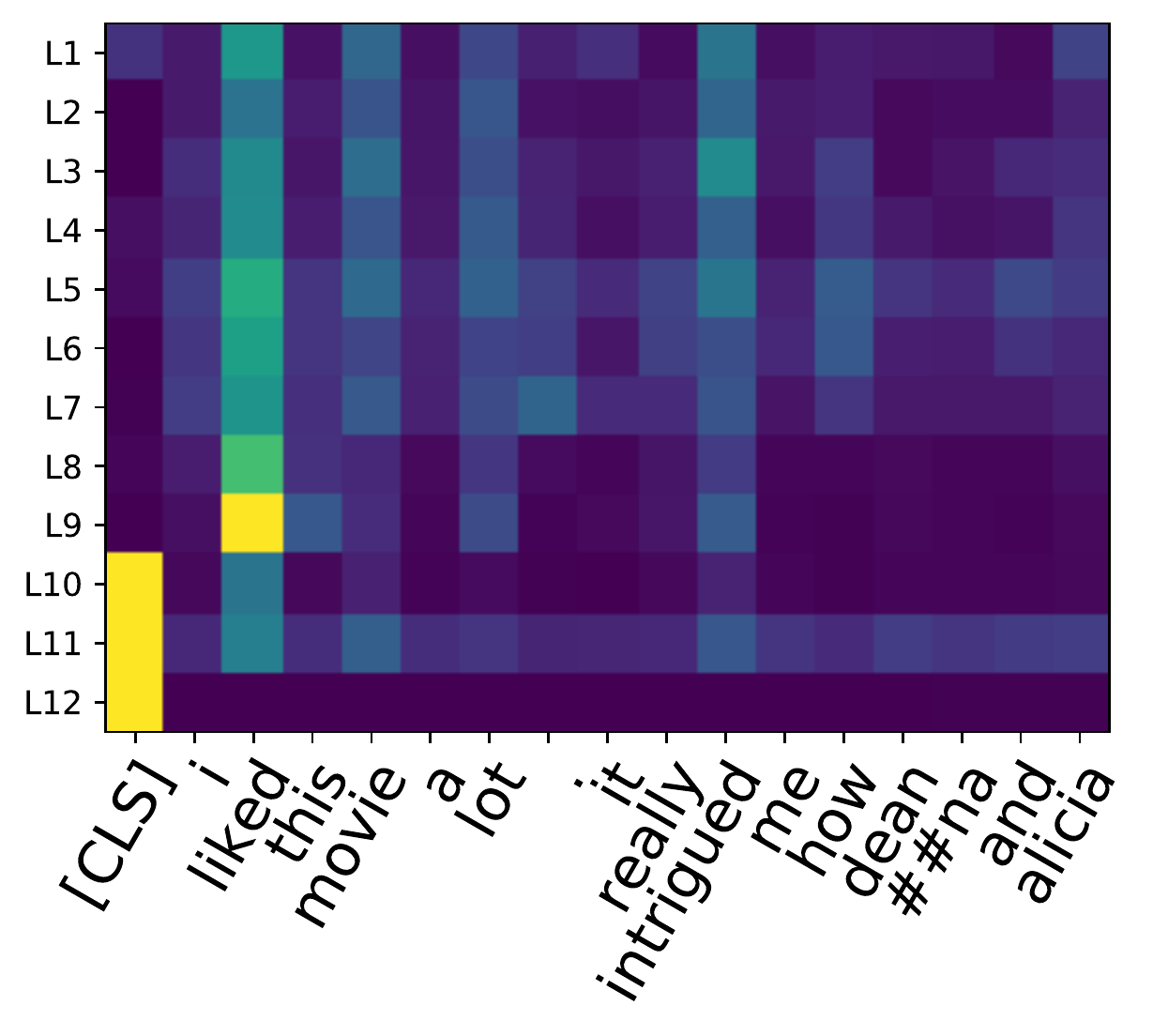}
        \label{subfig:before}
    \end{subfigure}%
    \begin{subfigure}{.35\textwidth}
        \centering
        \includegraphics[width=1\linewidth]{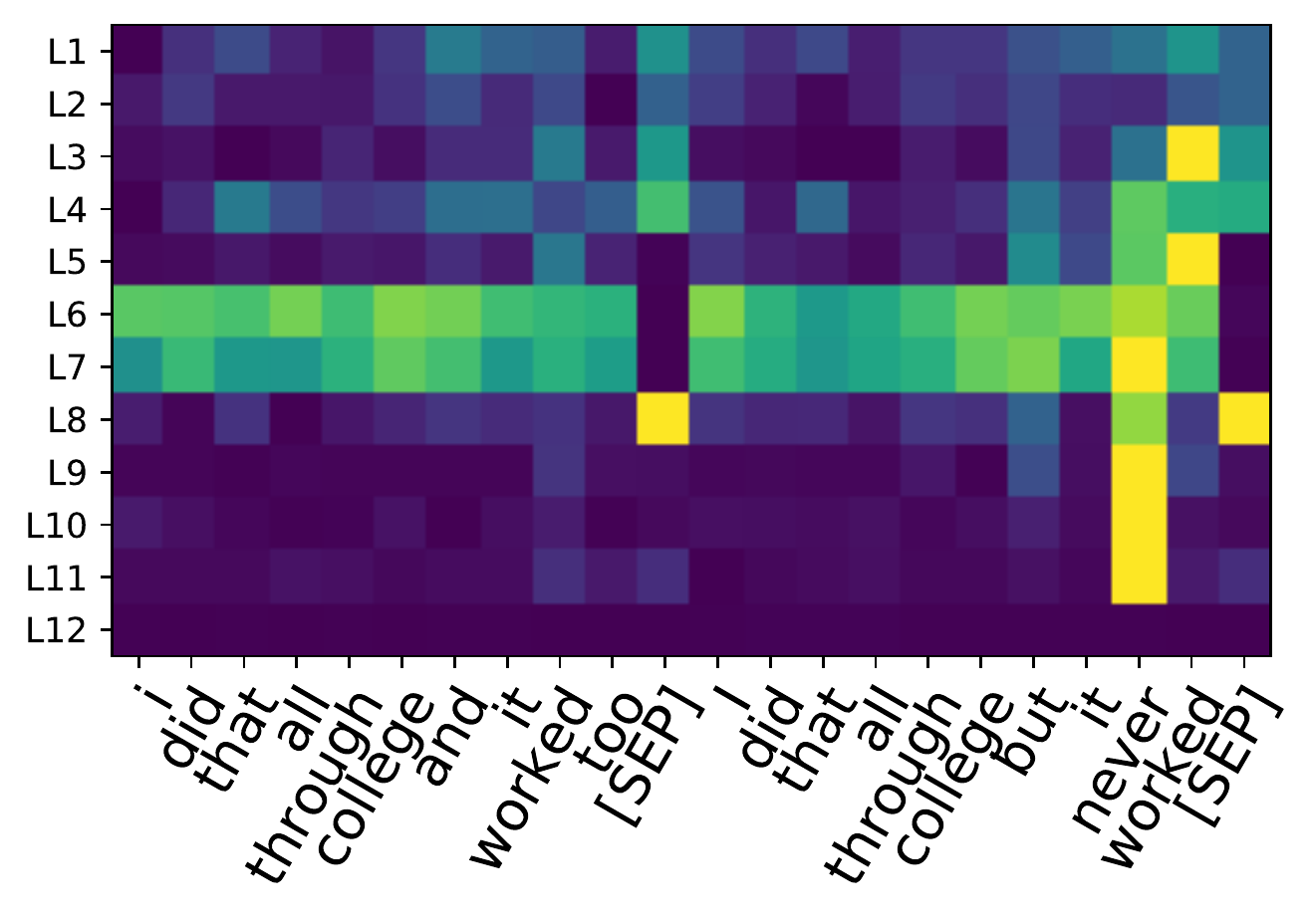}
        \label{subfig:half}
    \end{subfigure}
    \vspace{-20pt}
\caption{Illustrations from left to right are as follows: The before and after comparison of inserting an information bottleneck after layer 6; attribution for an IMDB example with the positive label; attribution for an MNLI example with the contradiction label.}
\label{fig:example}

\end{figure*}

\paragraph{Quantitative Analysis.}

Table \ref{tab:raw} shows the absolute probability drop $\|\bar{p}(y|x)-o\|$ with the first 11\% of the important tokens removed.
We further plot the normalized probability drop after each percentage of the important tokens is removed, as shown in Figure~\ref{fig:main}, indicating how much important information is lost for prediction: the steeper the slope, the better the ability to capture important tokens.
For this experiment, we insert the information bottleneck after layer 9, and we see that removing important tokens that are identified by our method deteriorates the probability the most on IMDB and MNLI Matched/Mismatched.

Of course, choosing the right layer to insert the information bottleneck is crucial to the result.
It also indicates which layer encodes the most meaningful information for prediction.
To investigate differences in inserting information bottlenecks after different layers, we carry the degradation test on 1000 random test samples across layers on IMDB, as shown in Figure \ref{subfig:layer}---see Appendix~\ref{C} for all 12 layers.
Insertion after layers 1, 8, and 9 generates more meaningful attribution scores.
At layer~1, the tokens remain distinct (i.e., representations have not been aggregated), and it is likely that the latent representation $\mathbf{T}$ is essentially capturing per-token sentiment values.
The big drop of $\bar{d}$ after layers 8 and 9, on the other hand, is interesting.
Recently, \citet{xin-etal-2020-deebert} examined early exit mechanisms in BERT and found that halting inference at layers 8 or 9 produces results not much worse than full inference, which suggests that an abundance of information is encoded in those layers. 

Another important parameter is $\beta$, which controls the trade-off between restricting the information flow and achieving greater accuracy.
A smaller $\beta$ allows more information through, and an extremely small $\beta$ has the same effect of using $\mathbf{X}$ as the attribution map.
As Figure \ref{subfig:beta} shows, when $\beta\leq 1e-6$, the degradation curve is similar to the one using $\mathbf{X}$ only. Appendix~\ref{E} shows the effects of different $\beta$ on a specific example.

\paragraph{Qualitative Analysis.}

The first plot in Figure~\ref{fig:example} shows the before and after comparison of IB insertion, with positive tokens highlighted. 
The second and third plots visualize attribution maps for instances across layers. 
Consistent with our quantitative analysis in Figure \ref{subfig:layer}, these plots demonstrate that, for a fully fine-tuned BERT, layers 8 and 9 seem to encode the most important information for the prediction.
For example, in the IMDB instance, \textit{liked} and \textit{intrigued} have the highest attribution scores for the prediction of positive sentiment across most layers---see layer 9 in particular.
In the MNLI example, \textit{never} is mostly highlighted starting from layer 7 to predict ``contradiction.'' 

\section{Conclusion}
In this paper, we adopt an information-bottleneck-based approach to analyze attribution for transformers. Our method outperforms two widely used attribution methods across four datasets in sentiment analysis, document classification, and textual entailment. We also analyze the information across layers both quantitatively and qualitatively.

\section*{Acknowledgments}

This research was supported in part by the Canada First Research Excellence Fund and the Natural Sciences and Engineering Research Council (NSERC) of Canada.

\bibliography{emnlp2020}
\bibliographystyle{acl_natbib}

\newpage
\appendix
\clearpage
\appendix
\section{Proof of Variational Upper Bound}
\label{B}
\begin{align*}
    \small
    \operatorname{I}(\mathbf{X}; \mathbf{T}) 
                                             & = \mathbb{E}_{\mathbf{X}}[D_{KL}[P(\mathbf{T} |\mathbf{X}) \| P(\mathbf{T})]] \label{eq:mi} \\
                                             & = \int_{\mathbf{X}}p(x)(\int_{\mathbf{T}}p(t|x)\log\frac{p(t|x)}{p(t)}dt)dx \\
                                             & = \int_{\mathbf{X}}\int_{\mathbf{T}}p(x, t)\log\frac{p(t|x)}{p(t)}\frac{q(t)}{q(t)} dtdx \\
                                             & = \int_{\mathbf{X}}\int_{\mathbf{T}}p(x, t)\log\frac{p(t|x)}{q(t)}dtdx \nonumber \\
                                             & + \int_{\mathbf{X}}\int_{\mathbf{T}}p(x, t)\log\frac{q(t)}{p(t)}dtdx \\
                                             & =  \int_{\mathbf{X}}\int_{\mathbf{T}}p(x, t)\log\frac{p(t|x)}{q(t)}dtdx \nonumber \\
                                             & + \int_{\mathbf{T}}p(t)(\int_{\mathbf{X}}p(x|t)dx)\log\frac{q(t)}{p(t)} dt \\
                                             & = \mathbb{E}_{\mathbf{X}}[D_{KL}[P(\mathbf{T}|\mathbf{X})\|Q(\mathbf{T})]] \nonumber \\
                                             & - D_{KL}[Q(\mathbf{T})\|P(\mathbf{T})] \\
                                             &\leq\mathbb{E}_{\mathbf{X}}[D_{KL}[P(\mathbf{T}|\mathbf{X})\|Q(\mathbf{T})]] 
\end{align*}

\vspace{0.25cm}

\section{Degradation Test across 12 Layers}
\label{C}

\begin{figure}[ht!]
    \centering
    \includegraphics[width=0.8\linewidth]{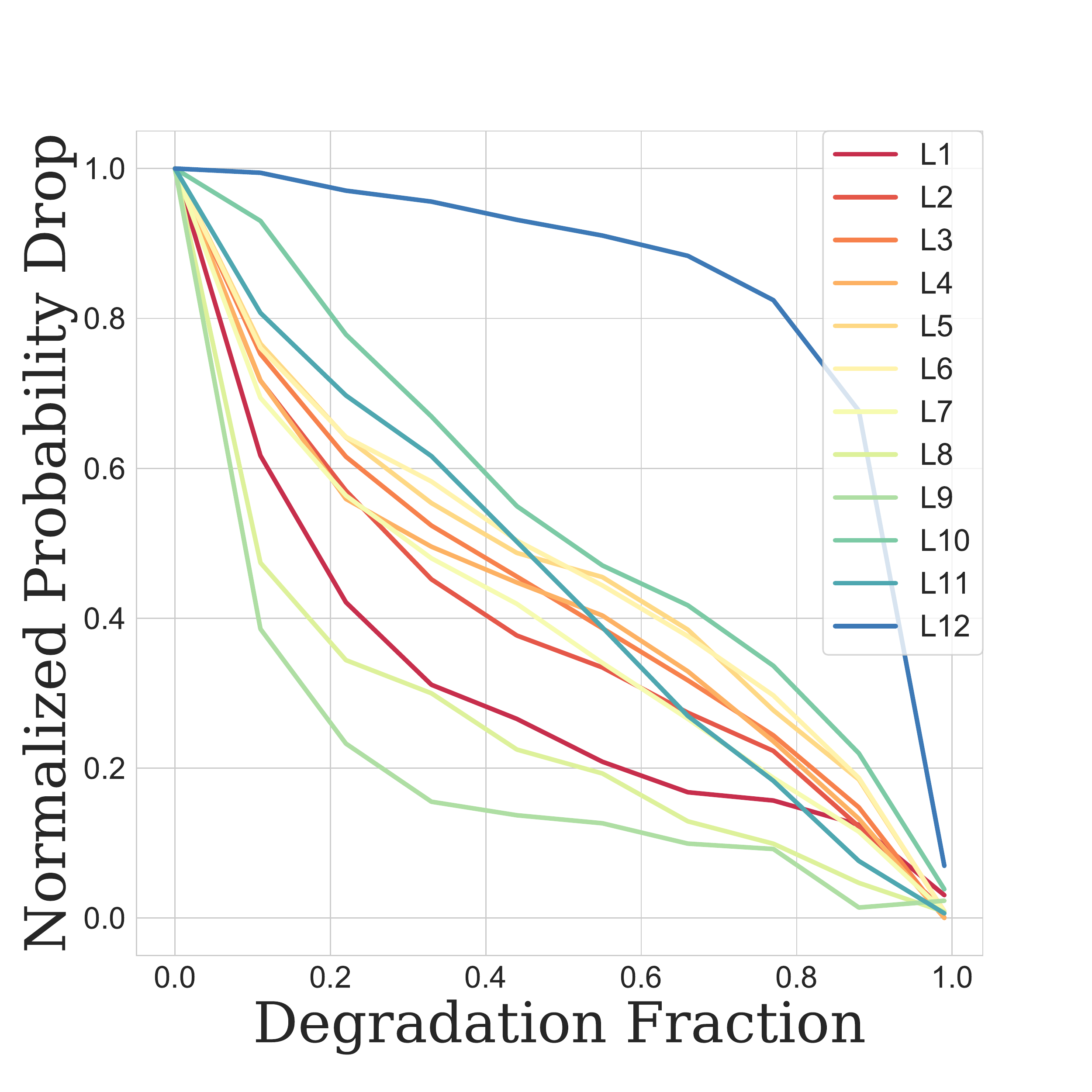}
    \caption{Degradation test results across all layers.}
    \label{fig:cross_layer}
\end{figure}

Figure \ref{fig:cross_layer} shows the complete version of the degradation test across all 12 layers. In general, the earlier we insert the bottleneck, the larger the probability drop is, except for layers 8 and 9, which are the only two layers with steeper slopes than layer 1.

\section{Visualization of the Effects of $\beta$}
\label{E}

\begin{figure}[ht!]
    \centering
    \begin{subfigure}{0.8\linewidth}
        \centering
        \includegraphics[width=1\linewidth]{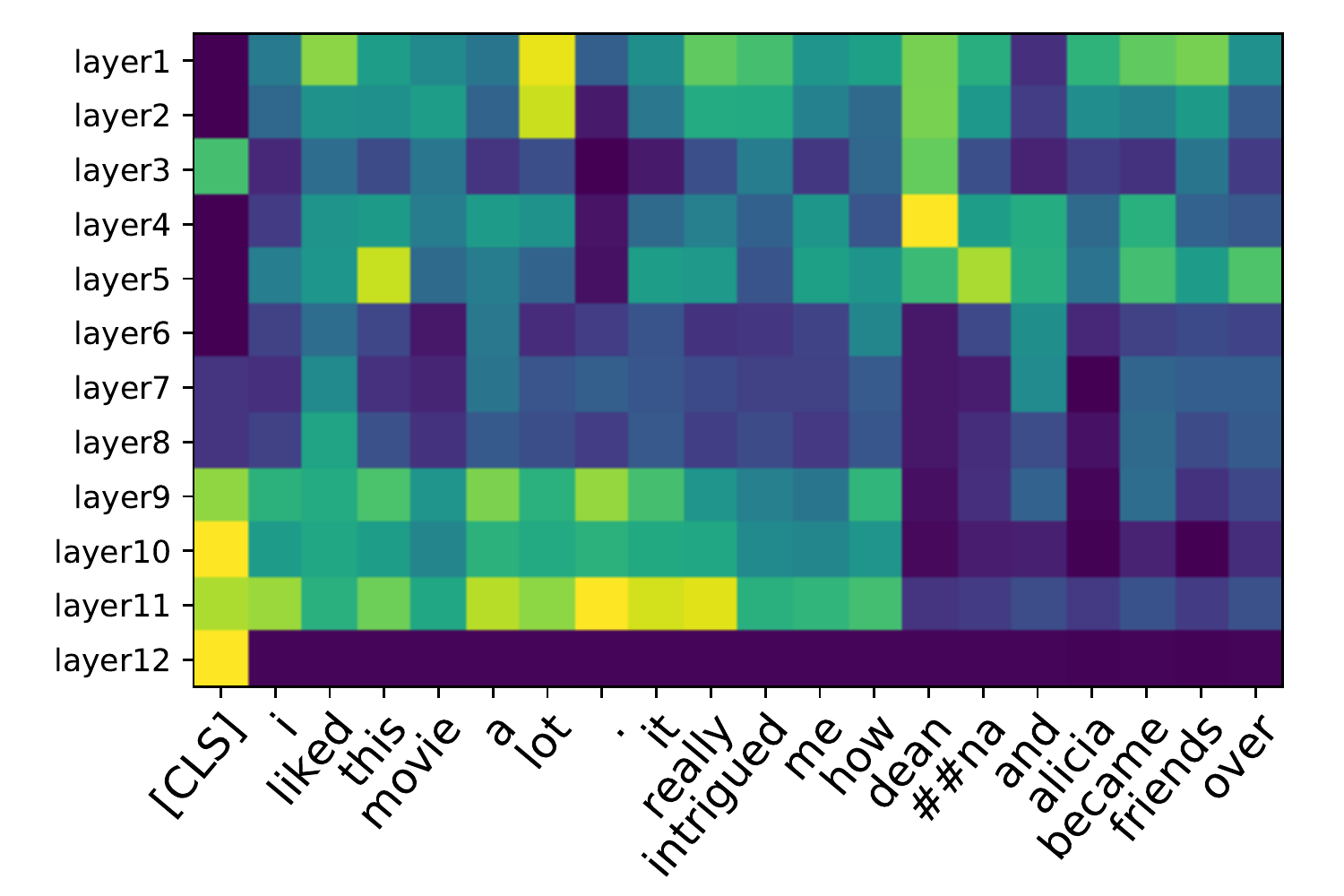}
        \caption{$\beta=10^{-3}$}
        \label{subfig:small_beta}
    \end{subfigure}
    
    \begin{subfigure}{0.8\linewidth}
        \centering
        \includegraphics[width=1\linewidth]{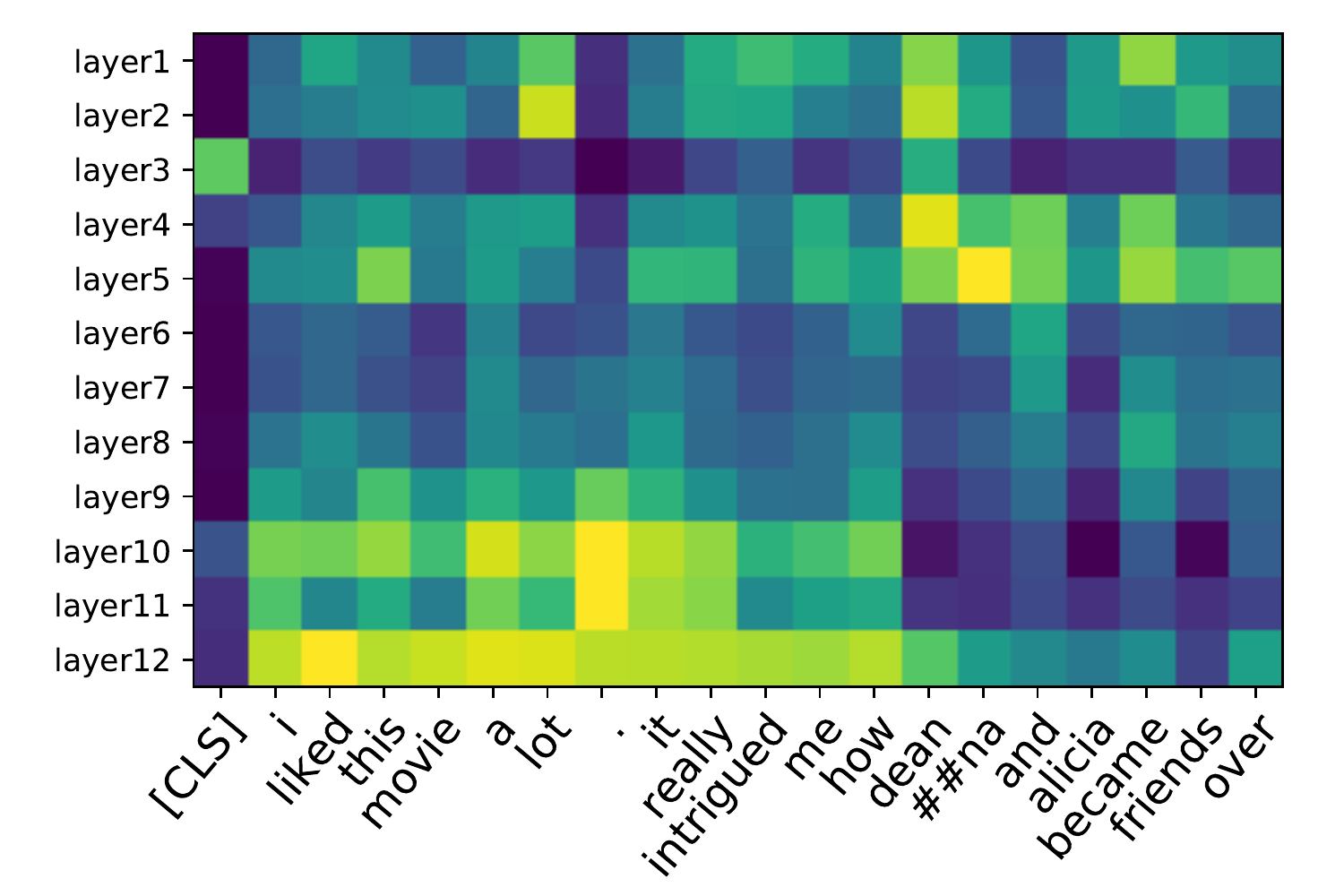}
        \caption{$\beta=10^{-7}$}
        \label{subfig:large_beta}
    \end{subfigure}
\caption{Comparison of BERT attribution maps with different values of $\beta$.}
\label{fig:vary_beta}

\end{figure}

Figure \ref{fig:vary_beta} shows the effects of different $\beta$ on a specific example. As we can see, when $\beta$ is as small as $10^{-7}$, most information is allowed to flow through the network and thus most parts are highlighted. In contrast, when $\beta$ is larger, the representation is more restricted.

\section{Detailed Parameters and Dataset Information}
\label{A}

To keep as much information as possible at the beginning, $\boldsymbol{\mu}_i$ should be set close to 1,$\forall i$, in which case $\mathbf{T}\approx \mathbf{X}$. So we initialize with $\alpha_i=5, \forall i$ and therefore $\boldsymbol{\mu}_i\approx 0.993$.
In order to stabilize the result, the input of the bottleneck ($\mathbf{X}$) is duplicated 10 times with different noise added.
We set the learning rate to 1 and the number of training steps to 10.
We use empirical estimation for $\beta\approx 10 \times \frac{\mathcal{L}_{CE}}{\mathcal{L}_{IB}}$. 
For IMDB, MNLI Matched/Mismatched, and AGNews, we insert the IB after layer 9 and $\beta$ is set to $10^{-5}$.
For RTE, we insert the IB after layer 10 and $\beta$ is set to $10^{-4}$. 

We carry out experiments on NVIDIA RTX 2080 Ti GPUs with 11GB VRAM running PyTorch 1.4.0 and CUDA 10.0.
A full technical description of our computing environment is released alongside our codebase.
For LIME, we set $N$, the number of permuted samples drawn from the original dataset, to 100 as this reaches the limitation of GPU memory.
Similarly, the number of steps of integrated gradients is set to 10 because it is more memory intensive.
The average time of running 25000 instances on the described GPU is about 10 hours for IBA, 13 hours for LIME, and 2 hours for IG.

\begin{table}[h]
    \centering
    \begin{tabular}{c|c}
        Dataset & Number of Dev/Test \\
        \toprule
        IMDB & 25000 \\
        MNLI Matched & 9815 \\
        MNLI Mismatched & 9832 \\
        AG News & 7600 \\
        RTE & 277 \\
        \bottomrule
    \end{tabular}
    \caption{Dataset Details.}
    \label{tab:data}
\end{table}

We use the test sets when the label is provided and use the dev sets otherwise. 
See Table~\ref{tab:data} for details. 
Note that ``IMDB'' refers to the sentiment analysis dataset provided by \citet{maas-EtAl:2011:ACL-HLT2011}.
``MNLI Matched'' means that the training set and the test set have the same set of genres while ``MNLI Mismatched'' means that genres that appear in the test set don't appear in the training set. 
Detailed information of the MNLI dataset can be found in \citet{williams-etal-2018-broad}.

\end{document}